\documentclass[conference]{IEEEtran}
	\input Macros/alphabet
	\input Macros/abrege
\def\pth#1{\left(#1\right)}                
              
\def\cro#1{\left[#1\right]}                
               
\def\norm#1{\left\|#1\right\|}

\def\FOR{\text{for\:}}

\newsavebox{\fminibox}
\newlength{\fminilength}
\newenvironment{fminipage}[1][\linewidth]
  {\setlength{\fminilength}{#1}
   \begin{lrbox}{\fminibox}\begin{minipage}{\fminilength}}
  {\end{minipage}\end{lrbox}\noindent\fbox{\usebox{\fminibox}}}

	\def\pmu{^{-1}}

   \def\wh#1{\widehat{#1}}                 

	\def\T{^\tD} 
	\def\I{\,|\,}           

   \def\froc#1#2{{#1/#2}}                

   \def\rond#1{\overset{\kern-0.33em~_\circ}{#1}}
   \def\rondit[#1]#2{\overset{\kern#1~_\circ}{#2}}

\def\babs{\begin{abstract}}             \def\eabs{\end{abstract}}
\def\barr{\begin{array}}                \def\earr{\end{array}}
\def\bcc{\begin{center}}                \def\ecc{\end{center}}
\def\bdes{\begin{description}}          \def\edes{\end{description}}
\def\bdoc{\begin{document}}             \def\edoc{\end{document}}
\def\ben{\begin{enumerate}}             \def\een{\end{enumerate}}
\def\barr{\begin{array}}                \def\earr{\end{array}}
\def\beqn{\begin{eqnarray}}             \def\eeqn{\end{eqnarray}}
\def\beqnx{\begin{eqnarray*}}           \def\eeqnx{\end{eqnarray*}}
\def\bseqn{\begin{subeqnarray}}         \def\eseqn{\end{subeqnarray}}
\def\beq#1\eeq{\begin{equation}#1   	\end{equation}}
\def\bal#1\eal{\begin{align}#1\end{align}}
\def\balx#1\ealx{\begin{align*}#1\end{align*}}
\def\beqx{$$}                           \def\eeqx{$$}
\def\bfig{\protect\begin{figure}}       \def\efig{\protect\end{figure}}
\def\bfigx{\protect\begin{figure*}}     \def\efigx{\protect\end{figure*}}
\def\bfigt{\protect\begin{figurette}}   \def\efigt{\protect\end{figurette}}
\def\bfl{\begin{flushleft}}             \def\efl{\end{flushleft}}
\def\bfr{\begin{flushright}}            \def\efr{\end{flushright}}
\def\bit{\begin{itemize}}               \def\eit{\end{itemize}}
\def\bmi{\begin{minipage}}              \def\emi{\end{minipage}}
\def\bfmi{\begin{fminipage}}            \def\efmi{\end{fminipage}}
\def\bpic{\begin{picture}}              \def\epic{\end{picture}}
\def\bqun{\begin{quotation}}            \def\equn{\end{quotation}}
\def\bsl{\begin{slide}}               	\def\esl{\end{slide}}
\def\btabb{\begin{tabbing}}             \def\etabb{\end{tabbing}}
\def\btabl{\begin{table}}               \def\etabl{\end{table}}
\def\btablx{\begin{table*}}             \def\etablx{\end{table*}}
\def\btabu{\begin{tabular}}             \def\etabu{\end{tabular}}
\def\btabx{\begin{tabular*}}            \def\etabx{\end{tabular*}}
\def\bbib{\begin{thebibliography}{999}} \def\ebib{\end{thebibliography}}
\def\bver{\begin{verbatim}}             \def\ever{\end{verbatim}}
\def\bca{\begin{cases}}                 \def\eca{\end{cases}}
\def\brk{\begin{remark}~}                \def\erk{\end{remark}}

\def\cl#1{\centerline{#1}}

    \usepackage{cite}
    \usepackage{amsmath,amssymb,amsfonts}
    \usepackage{graphicx}
    \usepackage{xcolor}
   	\usepackage{tikz}\usetikzlibrary{shapes.geometric,patterns,arrows}

	\definecolor{Bleu}{rgb}{0,0,0.75}    \def\Blue#1{{\color{Bleu} #1}}
	\definecolor{Rouge}{rgb}{0.75,0,0}   \def\Red#1{{\color{Rouge} #1}}
	\definecolor{Vert}{rgb}{0,0.7,0}     \def\Green#1{{\color{Vert} #1}}
    \definecolor{Marron}{rgb}{0.5,0.5,0} \def\Brown#1{{\color{Marron} #1}}
    \definecolor{Magenta}{rgb}{1,0,1}    \def\Magenta#1{{\color{Magenta} #1}}
    \definecolor{Gris}{rgb}{1,0,1}    	 \def\Gray#1{{\color{gray} #1}}

	\def\fns{\footnotesize}
	\def\sss{\scriptscriptstyle}
	\def\For{\xb}
	\def\Back{\xb}
    \def\VarFor{v^{\sss +}}		\def\SigFor{\sigma^{\sss +}}
    \def\VarBack{v^{\sss -}}	\def\SigBack{\sigma^{\sss -}}
	\def\Bruitb{\eb} \def\Bruit{e}
	\def\colog{\mathrm{colog}\,}
    \usepackage{pifont} \def\MyCheck{\ding{51}}
	\def\Quote#1{``#1''}

\begin{document}

\title{ A Gibbs posterior sampler \\ for inverse problem  based on prior diffusion model}
 
\author{
\IEEEauthorblockN{Jean-Fran\c{c}ois~Giovannelli}
\IEEEauthorblockA{\textit{Groupe Signal-Image, IMS (Univ. Bordeaux, CNRS, BINP), Talence, France} }
%
%
%
}

\maketitle

\begin{abstract}
This paper addresses the issue of inversion in cases where (1) the observation system is modeled by a linear transformation and additive error, (2) the problem is ill-posed and regularization relies on a Bayesian strategy, (3)~the prior is modeled by a diffusion process adjusted on an available large set of examples. In this context, it is known that the issue of posterior sampling is a thorny one and the paper introduces a Gibbs algorithm. It appears that this avenue has not been explored, and we show that it is particularly effective and remarkably simple. In addition, it provides clear elements regarding convergence guarantees in a specific case and arguments supporting such guarantees in practical cases. The results are clearly confirmed by numerical simulations based on a toy example.
\end{abstract}


\section{Introduction and problem formulation}\label{Sec:Problem}

The present paper is devoted to the resolution of inverse problems (\eg\cite{Giovannelli15,Kaipio05,Santamarina05,Hansen10} and references therein) when the observation system can be modelled by a linear operator and an additive Gaussian error:
\beq\label{Eq:ModelDirect}
\yb = \Hb \xb_{\sss 0} + \eb,
\eeq
where $\xb_{\sss 0}$ in $\eR^P$ collects the unknowns, $\yb$ and $\eb$ in $\eR^M$ collect the measurements and errors, and $\Hb$ in $\eR^P\times \eR^M$ characterizes the operator, \eg a convolution. 
The paper focuses on the ill-posed case and the regularization is provided within a Bayesian framework through a prior $\pi_{\sss 0}(\xb_{\sss 0})$. It is considered that a large example set is available, presumed to be made of samples from $\pi_{\sss 0}$. 

To describe this prior, a diffusion model is used (\eg\cite{Chan24,Ribeiro25,Nakkiran25} and references therein): in substance, examples are transformed into noise, and conversely, new examples are generated by transforming noise realizations. To do so, let us define $T$ latent variables $\xb_{\sss 1:T}$ (in addition to $\xb_{\sss 0}$) and an extended prior $\pi_{\sss 0:T}(\xb_{\sss 0:T})$. In order to define the latter, let us introduce two joint pdfs for $\xb_{\sss 0:T}$: a \textsl{forward} denoted $p^{\sss +}_{\sss 0:T}$ and a \textsl{backward} denoted $p^{\sss -}_{\sss 0:T}$. For practical efficiency, both are chosen in Markovian form:  
\beqn
p^{\sss +}_{\sss 0:T}(\For_{\sss 0:T})
&=& p^{\sss +}_{\sss 0} (\For_{\sss 0}) \, \prod\nolimits_{\sss t=1}^{\sss T} \, p^{\sss +}_{\sss t|t-1} (\For_{\sss t} \I \For_{\sss t-1}) \label{Eq:Forward} \\
p^{\sss -}_{\sss 0:T}(\Back_{\sss 0:T}) 
 &=& p^{\sss -}_{\sss T} (\Back_{\sss T}) \, \prod\nolimits_{\sss t=1}^{\sss T} p^{\sss -}_{\sss t-1|t} (\Back_{\sss t-1} \I \Back_{\sss t}) \label{Eq:Backward}
\eeqn
which reveals two terminal marginal pdfs $p^{\sss +}_{\sss 0}$ and $p^{\sss -}_{\sss T}$ and two sets of transition pdfs $p^{\sss +}_{\sss t|t-1}$ and $p^{\sss -}_{\sss t-1|t}$. 
Regarding the terminals
\beqnx
p^{\sss +}_{\sss 0} (\For_{\sss 0})  &=& \pi_{\sss 0}   (\For_{\sss 0})\\
p^{\sss -}_{\sss T} (\Back_{\sss T}) &=& \Nc( \Back_{\sss T} \,;\, \zerob , \Ib)
\eeqnx
the first is the pdf of the example set and the second is the pdf of noise (standard Gauss, \ie white, zero-mean, unitary variance). 
With regard to transitions, again for practical efficiency, Gaussians are chosen with the following parameters.
\beqn
p^{\sss +}_{\sss t|t-1} (\For_{\sss t}  \I \For_{\sss t-1})  &=& \Nc (\For_{\sss t}   \,;\, k_{\sss t} \, \For_{\sss t-1}    , \VarFor_{\sss t} \Ib  ) \label{Eq:ForwardTransit} \\
p^{\sss -}_{\sss t-1|t} (\Back_{\sss t-1} \I \Back_{\sss t}) &=& \Nc ( \Back_{\sss t-1} \,;\, \mub_{\sss t}(\Back_{\sss t})  , \VarBack_{\sss t} \Ib ) \label{Eq:BackwardTransit}
\eeqn
The function $\mub_{\sss t}$ can be seen as a \textsl{denoiser} and it is described by a neural network $\mub_{\sss t}^{\sss\thetab}$, with parameter $\thetab$. It has two inputs: the image $\Back$ (to be denoised) and the time $t$ providing information on the levels of noise  $\VarFor_{\sss t}$ and attenuation  $k_{\sss t}$. Replacing $\mub_{\sss t}$ by $\mub^{\sss\thetab}_{\sss t}$ in (\ref{Eq:BackwardTransit}), and substituting in (\ref{Eq:Backward}), yields the $p^{\sss -,\thetab}_{\sss t|t+1}$ then $p^{\sss -,\thetab}_{\sss 0:T}$.
Regarding the parameters $k_{\sss t}$, $\VarFor_{\sss t}$, $\VarBack_{\sss t}$ we refer to the usual Variance Preserving scheme~\cite{Chan24,Ribeiro25,Nakkiran25}. 
%

The learning stage adjusts $\thetab$ to minimise the Kullback distance between forward $p^{\sss +}_{\sss 0:T}$ and backward $p^{\sss -,\thetab}_{\sss 0:T}$ joint pdfs
\beqx\label{Eq:Kullback}
	\Dc \cro{\, p^{\sss +}_{\sss 0:T} \,;\, p^{\sss -,\thetab}_{\sss 0:T} \,}
\eeqx
while ensuring that the marginal pdfs for $\xb_{\sss 0}$ and $\xb_{\sss T}$ are
\beqx
\bca
    \pi_{\sss 0}	& \simeq~ p^{\sss +}_{\sss 0} ~\simeq~ p^{\sss -}_{\sss 0} \cr
    \Nc				& \simeq~ p^{\sss -}_{\sss T} ~\simeq~ p^{\sss +}_{\sss T} \cr
\eca
\eeqx	
\ie that of the example set and the noise. 
It suffices then to report the adjusted value of $\thetab$ in $\mub_{\sss t}^{\sss\thetab}(\Back)$ and $p^{\sss -,\thetab}_{\sss 0:T}$ to obtain an adjusted joint backward pdf~(\ref{Eq:Backward}). Based on its Markov and Gauss structure, it is then easy to sample: 
\beqnx
%
%
\xb_{\sss T}	&\sim& \Nc( \,\cdot\, \,;\, \zerob , \Ib)\\  
\xb_{\sss t-1}	&\sim& \Nc ( \,\cdot\, \,;\, \mub_{\sss t}^{\sss\thetab}(\xb_{\sss t}), \VarBack_{\sss t} \Ib ) \,, ~~\FOR t=T,\dots 2, 1 \,.
\eeqnx
Start by drawing an image $\xb_{\sss T}$ as a noise. Then, repeat the following two steps, from $t=T$ to $t=1$: 
\bit
\item pass the image through the network $\mub_{\sss t}^{\sss\thetab}(\xb_{\sss t})$ and 
\item add white noise with zero-mean and variance $\VarBack_{\sss t}$.
\eit
%
%
It is referred to as \textsl{ancestral sampling}. The results are particularly convincing and it is now a standard practice to seek to use this pdf and sampler in the Bayesian inversion. 

\begin{figure*}[htbp]
\bcc
\begin{tikzpicture}[>=triangle 45,xscale=1,yscale=1,rotate=270]
\node{$\yb$} [grow=up]  
child{ node {$\xb_{\sss 0}$} 
child{ node {$\xb_{\sss 1}$} 
child{ node {$\xb_{\sss 2}$} 
child{ node {$\cdots$} 
child{ node {$\xb_{\sss t-1}$} 
child{ node {$\xb_{\sss t}$} 
child{ node {$\xb_{\sss t+1}$} 
child{ node {$\cdots$} 
child{ node {$\xb_{\sss T-1}$} 
child{ node {$\xb_{\sss T}$} 
}}}}}}}}} };
\end{tikzpicture}
\ecc
\caption{Hierarchy: $\xb_{\sss 0}$ is the image of interest, $\xb_{\sss 1:T}$ are  the latent images and $\yb$ is the measured image (blurred and noisy version of the true $\xb_{\sss 0}$). It is, fundamentally, the keystone of the proposed Gibbs algorithm. \label{Fig:Hierarchy}} 
\end{figure*}

To do so, the measurements are included \via the likelihood deduced from the observation model~(\ref{Eq:ModelDirect}) and a model for the error. Here the latter is described as zero-mean and white with variance $v_\Bruit$: $f(\yb|\xb_{\sss 0}) = \Nc(\yb \,;\, \Hb\xb_{\sss 0},v_\Bruit \Ib$). 
We can then construct  the extended joint pdf $f(\yb,\xb_{\sss 0:T})=f(\yb|\xb_{\sss 0})\,\pi_{\sss 0:T}(\xb_{\sss 0:T})$ and the extended posterior: $\pi_{\sss 0:T}(\xb_{\sss 0:T}|\yb)  = \froc{ f(\yb,\xb_{\sss 0:T}) }{f(\yb) }$.
%
This development is based on a model with a natural conditional independence property: the pdfs for $(\yb|\xb_{\sss 0})$ and $(\yb|\xb_{\sss 0:T})$ are equal (see the hierarchical representation in Fig.\,\ref{Fig:Hierarchy}).

\section{Sampling: existing approaches and limitations} \label{Sec:Sampling}\label{Sec:Existing}

In the specific context of a prior based on a diffusion model, the main difficulty is to sample from the posterior. 

In recent years, there have been a number of papers devoted to this issue.
\cite{Chung24} and \cite{Song23} respectively introduce Diffusion Posterior Sampling (DPS) and Pseudoinverse-guided Diffusion Models ($\Pi$-GDM). They rely on the likelihood attached to $\xb_{\sss t}$ (instead of $\xb_{\sss 0}$) and on an approximation based on Tweedie's formula. \cite{Wu23} introduces the Twisted Diffusion Sampler (TDS): a SMC (Sequential Monte Carlo) algorithm targets the posterior, based on a set of weighted particles. \cite{Dou24} proposes an idea inherited from Bayesian filtering: Filtering Posterior Sampling (FPS) that introduces latent measurements. An extended version also leverages SMC methods. \cite{Zhu24} proposes Diffusion Posterior MCMC (DPMC) based on Annealed MCMC, \ie defines a series of intermediate pdfs. \cite{Cardoso23} defines a sequence of intermediate linear inverse problems, then also a sequence of pdfs that are sampled from by means of SMC. \cite{Coeurdoux24} introduces variable splitting to divide the sampling task into two simpler steps: a linear-Gauss one (Wiener\,/\,Tikhonov) and a denoising one carried out by the diffusion model.
%
%
\cite{Janati25} offers a comprehensive overview of MC methods and an analysis of the idea of twisting in relation with other approaches. 

Some methods are resolutely approximate~\cite{Chung24,Song23}, while others aim to target the posterior~\cite{Wu23,Dou24,Zhu24,Cardoso23}. The latter are essentially based on the concept of SMC and their convergence is for large enough number of particles.
In any case, they all appear to be inspired by the idea of ancestral sampling, in that each image $\xb_{\sss t}$ is visited once, and in the direction from $t=T$ to $t=0$. However, while ancestral sampling is certainly attractive for the prior, it is not effective for the posterior.

\section{Proposed sampler: a Gibbs scheme} \label{Sec:Proposed}\label{Sec:MCMC}

The paper now presents the contribution: a Gibbs algorithm for the joint posterior $\pi_{\sss 0:T}(\xb_{\sss 0:T}|\yb)$. It samples each $\xb_{\sss t}$ in turn\footnote{This is actually a Block-Gibbs version (each image $\xb_{\sss t}$ representing a block), but we will simply refer to it as Gibbs for the sake of simplicity.}, under its conditional pdf
%
$\pi_{\sss t|\star}(\xb_{\sss t}|\yb,\xb_{\sss \star\backslash t})$
%
where $(t\I\star)$ is the time $t$ given all the other times (from $0$ to $T$) except $t$ and $(\star\backslash t)$ denotes the set of all times (from $0$ to $T$) except $t$. 

The approach taken here aims to obtain a direct algorithm, in the sense that these conditionals can be sampled directly, typically a Gauss. 
The original idea is to play with both forward and backward pdfs (switch to play, say). More specifically, the sampling is based on the posterior attached to the 
\bit

\item forward $\pi^{\sss +}_{\sss 0:T}(\xb_{\sss 0:T}|\yb)$ for the latent variables $\xb_{\sss 1:T}$, and 

\item backward $\pi^{\sss -}_{\sss 0:T}(\xb_{\sss 0:T}|\yb)$ for the object of interest $\xb_{\sss 0}$.

\eit
This idea is justified by the fact that the two joint priors $\pi^{\sss +}_{\sss 0:T}(\xb_{\sss 0:T})$ and $\pi^{\sss -}_{\sss 0:T}(\xb_{\sss 0:T})$ are similar thanks to the learning stage, and we consider here they are identical. Work is ongoing to account for the fact that they are not strictly identical.

\medskip\indent\textit{1~-~Image of interest}~---~
Regarding $\xb_{\sss 0}$, it is sampled under $\pi_{\sss 0|\star}(\xb_{\sss 0}|\yb,\xb_{\sss \star\backslash 0})$ and given the hierarchy (Fig.\,\ref{Fig:Hierarchy})
\begin{align*}
\pi_{\sss 0|\star}(\xb_{\sss 0}|\yb,\xb_{\sss \star\backslash 0}) 
	& \propto~  \pi^{\sss -}_{\sss 0|1} (\xb_{\sss 0} \I \xb_{\sss 1}) ~ f(\yb \I \xb_{0})  \\
	& =~ \Nc( \xb_{\sss 0} \,; \mub_{\sss 1}(\xb_{\sss 1}) , \VarBack_{\sss 1} \Ib ) ~ \Nc(\yb \,;\, \Hb\xb_{\sss 0},v_\Bruit \Ib)
\end{align*}
%
%
%
which reveals a linear-Gauss problem.
The co-logarithm\footnote{to within one term and a factor of $2$\label{Foot:CoLog}} reads
\begin{align*}
\colog \pi_{\sss 0|\star}&(\xb_{\sss 0}|\yb,\xb_{\sss \star\backslash 0}) \\
	& \#~ \norm{\xb_{\sss 0} - \mub_{\sss 1}(\xb_{\sss 1}) }^2 / \VarBack_{\sss 1}  \,+\, \norm{\yb - \Hb\xb_{\sss 0}}^2 / v_\Bruit \,.
\end{align*}
So, the conditional posterior is Gauss with precision and expectation (involving the classic Wiener\,/\,Tikhonov solution).
\beqx
\bca
    \Gammab_{\sss 0} &=~~ \Hb\T \Hb \,/\, v_\Bruit  +  \Ib \,/\,\VarBack_{\sss 1}  \\
    \varepsilonb_{\sss 0} &=~~ \Gammab_{\sss 0}\pmu\pth{ \Hb\T \yb \,/\, v_\Bruit \,+\,  \mub_{\sss 1}(\xb_{1})\,/\,\VarBack_{\sss 1}}
\eca
\eeqx
Sampling is particularly effective in the Fourier plane: the components are independent and Gaussian, and their mean and variance are easily computed by simple FFT.

\medskip\indent\textit{2.1~-~Latent images ($t\neq T$)}~---~The case of $\xb_{\sss 1:T}$ also rely on the hierarchy (Fig.\,\ref{Fig:Hierarchy}). Naturally, it is needed to separate the termination $t=T$ from the other $t$. For $t=1,\dots, T-1$
\begin{align*}
\pi_{\sss t|\star}(\xb_{\sss t}|\yb,&\xb_{\sss \star\backslash t}) 
	 \propto~ \pi^{\sss +}_{\sss t|t-1} (\xb_{\sss t} \I \For_{\sss t-1}) ~ \pi^{\sss +}_{\sss t+1|t} (\xb_{\sss t+1} \I \xb_{\sss t})	\\
	& =~ \Nc (\xb_{\sss t}   \,;\, k_{\sss t} \, \xb_{\sss t-1} , \VarFor_{\sss t} \Ib  ) ~ \Nc (\xb_{\sss t+1}   \,;\, k_{\sss t+1} \, \xb_{\sss t} , \VarFor_{\sss t+1} \Ib  )
\end{align*}
%
%
so, by writing the co-logarithm\textsuperscript{\ref{Foot:CoLog}}
\begin{align*}
\colog & \pi_{\sss t|\star}(\xb_{\sss t}|\yb,\xb_{\sss \star\backslash t}) \\
	& \#~ \norm{\xb_{\sss t}  -  k_{\sss t} \, \xb_{\sss t-1} }^2 / \VarFor_{\sss t}  \,+\, \norm{\xb_{\sss t+1} - k_{\sss t+1} \, \xb_{\sss t} }^2 / \VarFor_{\sss t+1}
\end{align*}
a Gauss pdf is identified with precision $\gamma_{\sss t} \Ib$ and expectation~$\varepsilonb_{\sss t}$
\beqx
\bca
    \gamma_{\sss t} &=~~ 1 / \VarFor_{\sss t} \,+\,  k_{\sss t+1}^2 / \VarFor_{\sss t+1}  \\
    \varepsilonb_{\sss t} &=~~  \gamma_{\sss t}\pmu \pth{ k_{\sss t} \xb_{\sss t-1} / \VarFor_{\sss t} \,+\, k_{\sss t+1} \xb_{\sss t+1} / \VarFor_{\sss t+1} }
\eca
\eeqx

\medskip\indent\textit{2.2~-~Latent image ($t=T$)}~---~For the case of $\xb_{\sss T}$, based on the hierarchy (Fig.\,\ref{Fig:Hierarchy}):
\begin{align*}
\pi_{\sss T|\star}(\xb_{\sss T}|\yb,\xb_{\sss \star\backslash T}) 
	& =~ \pi^{\sss +}_{\sss T|T-1} (\For_{\sss T} \I \For_{\sss T-1}) \\
	& =~ \Nc (\For_{\sss T}   \,;\, k_{\sss T} \, \For_{\sss T-1} , \VarFor_{\sss T} \Ib  )
\end{align*}
\ie simply the last step in the forward process: a Gaussian with precision $\gamma_{\sss T} \Ib$ and expectation~$\varepsilonb_{\sss T}$%
\beqx
\bca
    \gamma_{\sss T} &=~~ 1/ \VarFor_{\sss T}   \\
    \varepsilonb_{\sss T} &=~~  k_{\sss T} \, \For_{\sss T-1}
\eca
\eeqx

\pagebreak
Overall, the entire algorithm is both simple and efficient.
\bit 
\item First, this requires the sampling of Gaussians only. \nocite{Kamilov25} 
\item Second, all the covariances are diagonal be it in the Fourier domain ($t=0$) or in the spatial one ($t\neq0$).
\item In addition, means and variances are easy to compute, by FFT ($t=0$) or linear combination ($t\neq0$). 
\eit
%
It is called G-DPS for Gibbs-Diffusion Posterior Sampling.

%
%
%
%

\textsl{Convergence property}~---~A crucial point is that, generally speaking, the Gibbs scheme is guaranteed to converge (for large enough number of iterations): it provides samples under a joint pdf by sampling the conditional pdfs (deduced form that joint pdf). However, this guaranty does not apply here since some conditionals are deduced from the forward joint pdf $\pi^{\sss +}_{\sss 0:T}$ and others from the backward joint pdf $\pi^{\sss -}_{\sss 0:T}$. That said, by its nature, the learning stage brings the forward and backward pdfs close together, supporting such convergence in practical cases. Furthermore, in a scenario where the two densities were equal, convergence would be rigorously guaranteed. 
%
%
%
%
%

\section{Numerical assessment}\label{Sec:Simulations}

\subsection{Simulation details}

An experimental study is proposed in order to demonstrate the feasibility and interest of the proposed G-DPS. It relies on a toy problem based on the MNIST example set. The method has been implemented\footnote{Computing environment Matlab on a standard PC with a 3.8~GHz CPU and 32~GB of RAM, with a GPU "NVIDIA GeForce RTX 3090".} and the information regarding the architecture and learning stage are given in~\cite{Matlab23}. For a first experiment: the ground-truth $\xb^\star$ is a sample of the learned prior (size $32\times 32$, gray level roughly in~$[0,1]$), the PSF is a $3 \times 3$ square with coefficient $1/9$, and the error level is  $\sigma_e=0.05$. 
The ground-truth $\xb^\star$ and the measurement $\yb$, \ie blurred and noisy image, are shown in Fig.\,\ref{Fig:ResultsImages} (left and middle).

Here are some points regarding implementation details.
\bit

\item The image $\xb_{0}$ is initialized to $\yb$. The $\xb_{\sss 1:T}$ are set to successive noisy versions through the forward scheme. 

\item Regarding the scan order: the algorithm update the images for $t=0, 1,\dots$ up to $t=T$ and repeats this pattern. 

\item As iterations proceed, the empirical average for $\xb_{\sss 0}$ is updated. The algorithm stops when the difference between successive updates is smaller than a threshold. 
%
	 
\eit 

\textsl{Remark}~---~The algorithm  has been run numerous times, under identical and different scenarios (ground-truth, error level, PSF and initializations): it has always shown consistent qualitative and quantitative behaviours and we have never detected any convergence or numerical instability issues.

\subsection{Results}

\bfig[h]
\centering
\includegraphics[height=6cm]{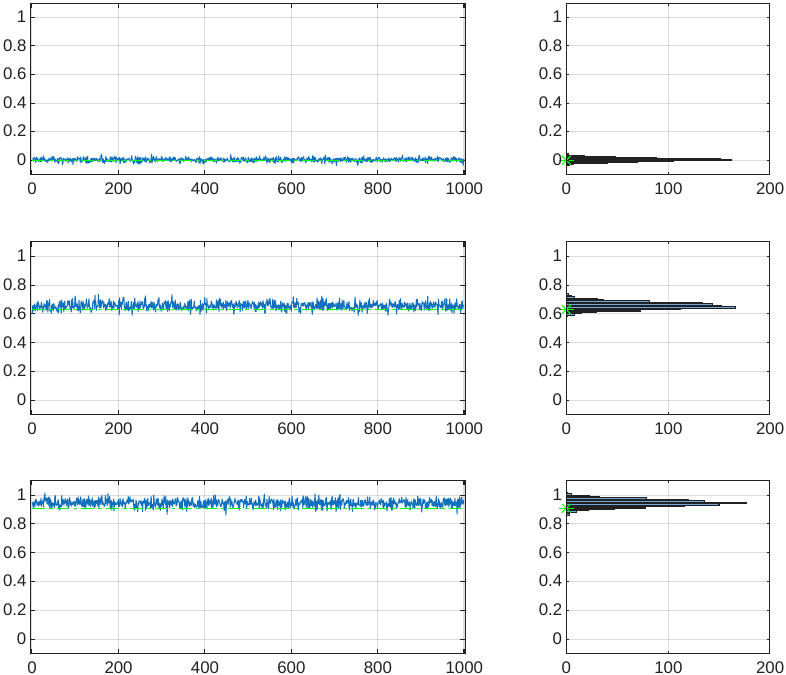}
\caption{Samples provided by the Gibbs algorithm for three pixels of $\xb_{\sss 0}$. They are shown as a function of iteration index (left) and as histograms (right). They are samples of one dimensional marginal pdfs. The green lines\,/\,dots give the true value. See also Fig.\,\ref{Fig:ResultsClouds} for two dimensional marginals pdfs and Tab.\,\ref{Tab:ParamEstimate} for quantitative assessment. \label{Fig:ResultsChains}}
\efig

\bfig[h]
\centering
\includegraphics[width=0.45\textwidth]{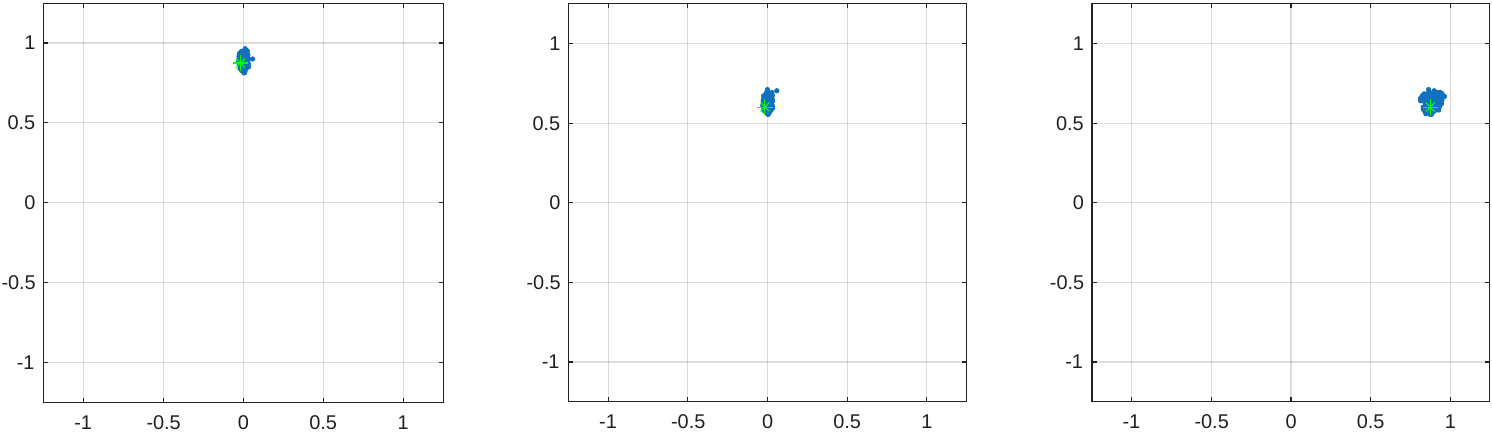} 
\caption{Point clouds for two dimensional marginals pdfs for three pixels. The samples are given in blue and the true value is given in green. See also Fig.\,\ref{Fig:ResultsChains} for one dimensional plots and Tab.\,\ref{Tab:ParamEstimate} for quantitative assessment. \label{Fig:ResultsClouds}}
\efig

Fig.\,\ref{Fig:ResultsChains} shows a typical result regarding three pixels. A standard behaviour is observed for the chains: the distributions quickly stabilise and seem stationary after about only $10$ iterations (burn-in period). 
%
%
From a qualitative view, Fig.\,\ref{Fig:ResultsChains} shows that the estimated gray levels are nearby the ground-truth (known in this simulation study). The quantitative results are reported in Tab.\,\ref{Tab:ParamEstimate}. 



\btabl[htb]
\bcc
\btabu{lrrr}
            & Pixel\,1 &  Pixel\,2    &  Pixel\,3 \\[0.01cm] \hline
True        & -0.00142 & 0.55092      & 0.61704   \\
Estimate    &  0.00122 & 0.61773      & 0.69163   \\
Error       &  0.0026\phantom{0} & 0.0668\phantom{0}      & 0.0746\phantom{0}    \\
PSD         &  0.01020 & 0.05730      & 0.05719     \\
$\pm$ 2PSD  & \MyCheck~~~ & \MyCheck~~~     & \MyCheck~~~  \\
\etabu
\ecc
\caption{Results for three pixels: true and estimated values (first and second row) then the error (third row). The Posterior Standard Deviation (PSD) is then given and the \MyCheck\, indicates that the true value does lie within the interval with center the estimate and width two PSD. See also Figs.\,\ref{Fig:ResultsChains} and~\ref{Fig:ResultsClouds}. \label{Tab:ParamEstimate}}
\etabl

Finally, Fig.\,\ref{Fig:ResultsImages}-right yield the estimated image. The blur and the noise are significantly reduced in the resulting image (Fig.\,\ref{Fig:ResultsImages}-right) with respect to the measurement (Fig.\,\ref{Fig:ResultsImages}-middle) and it closely matches the original image (Fig.\,\ref{Fig:ResultsImages}-left). This result is confirmed by the cross-sections also in presented in Fig.\,\ref{Fig:ResultsImages} and by the quantitative results in Tab.\,\ref{Tab:ParamEstimate}.

\bfig[htb]
\btabu{rrr}
\includegraphics[width=2.3cm]{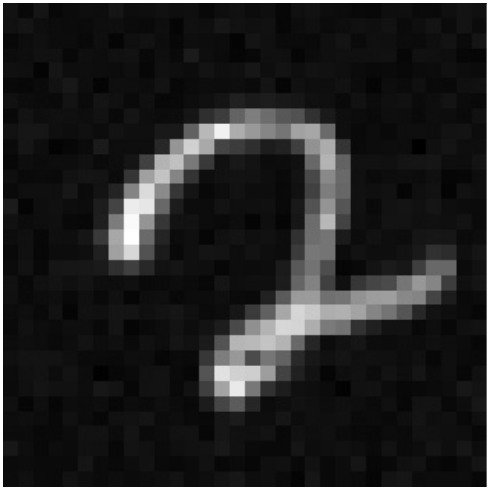} &
\includegraphics[width=2.3cm]{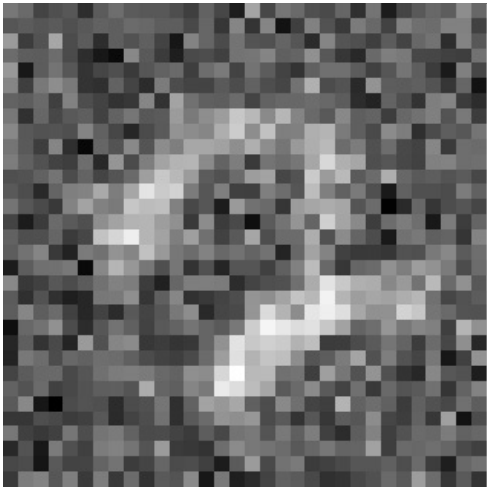}  &
\includegraphics[width=2.3cm]{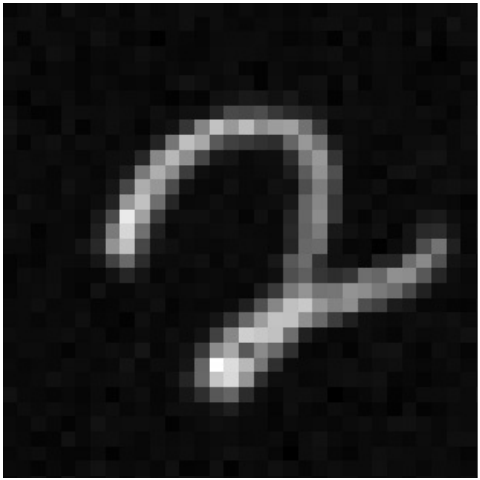}   \\
\includegraphics[width=2.4cm]{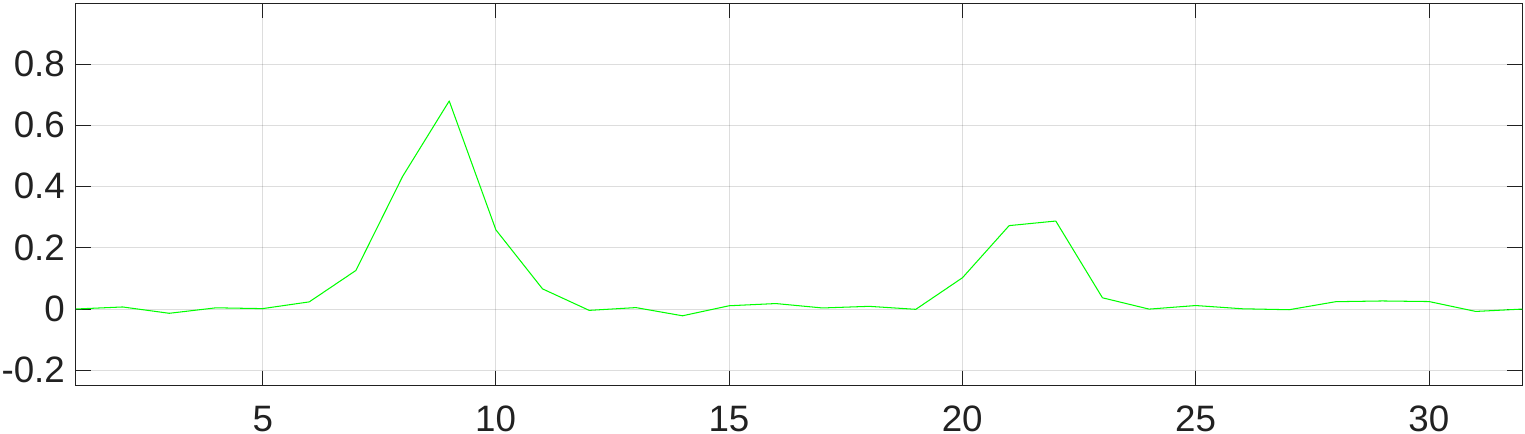} &
\includegraphics[width=2.4cm]{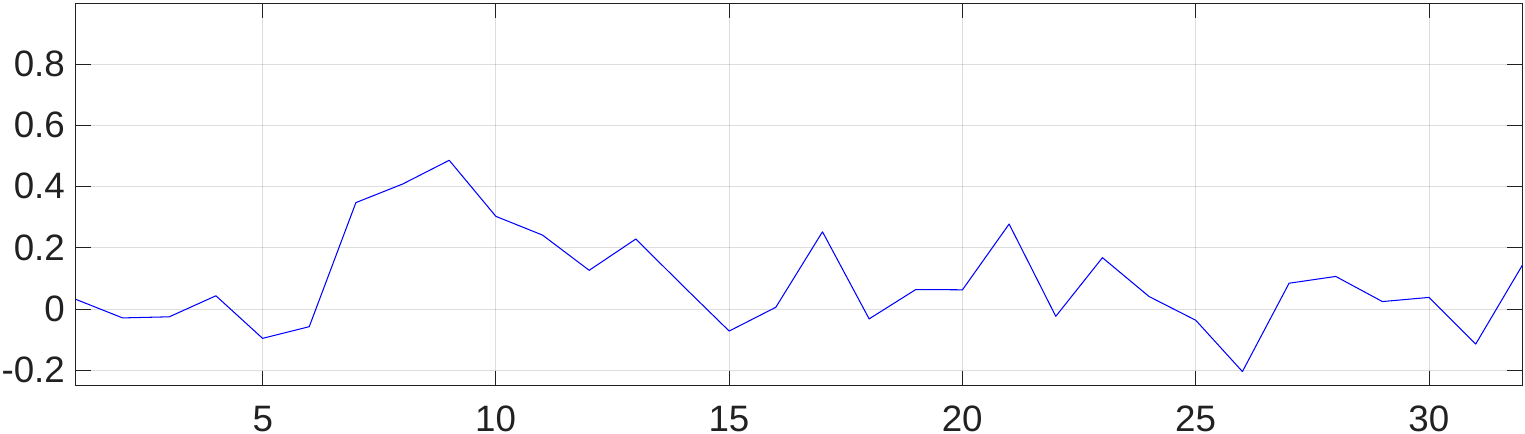}  &
\includegraphics[width=2.4cm]{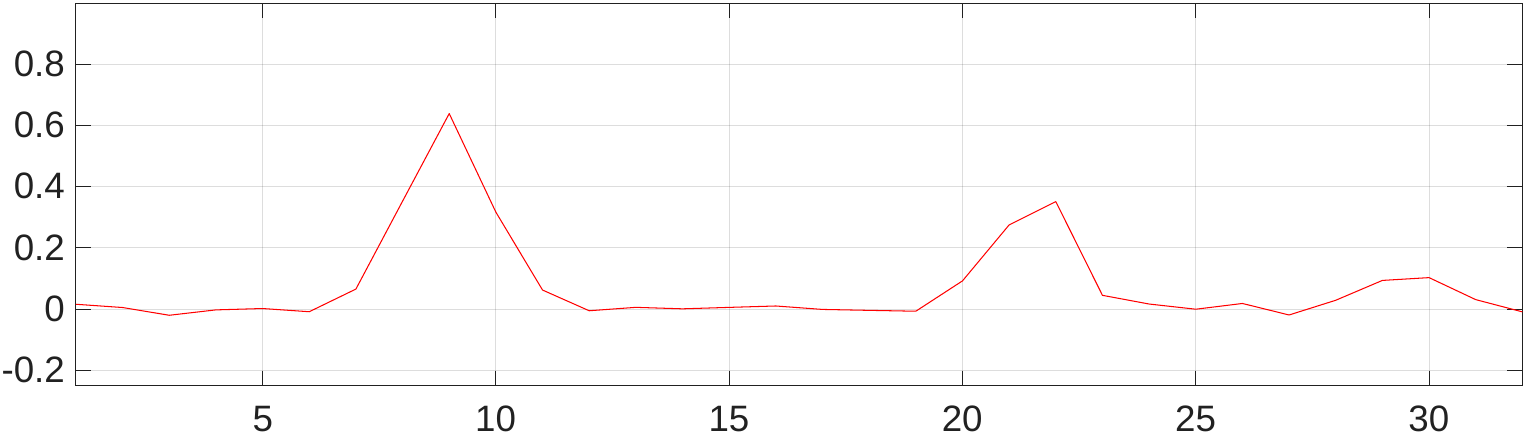}   \\
\etabu
\caption{Left to right: true object $\xb^\star$, measurements $\yb$ and estimated object $\wh\xb$. The figure shows the images themselves (top) and cross-sections (bottom). \label{Fig:ResultsImages}}
\efig

%
As a Bayesian and sampling one, the proposed strategy provides not only optimal estimation for the unknowns (\eg Posterior Mean as the MMSE), but also coherent tool for uncertainty quantification. More specifically, sampling the posterior enable to compute (a) estimation and (b) posterior standard deviations (PSD) as empirical averages based on the samples. 
From Fig.\,\ref{Fig:ResultsUncertainty} and Tab.\,\ref{Tab:ParamEstimate} it is clear that, for each pixel, the true value lies within the interval of center the estimate and of width two standard deviations. This reinforces the relevance of the Bayesian strategy and the importance of the quality of posterior sampling for the correct quantification of uncertainty.
%

\bfig[htb]
\bcc
\includegraphics[width=7cm]{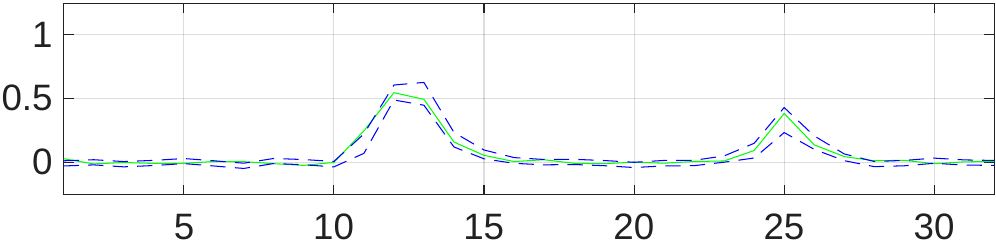}   \\
\ecc
\caption{Cross-sections of $\xb^\star$ (plain green) and the \Quote{uncertainty} intervals (dashed blue) with center the estimate and of width two standard deviations.} \label{Fig:ResultsUncertainty}
\efig

\subsection{Additional consideration regarding convergence speed} \label{Sec:SimulationsAdditional}

This section explores the topic of speed of convergence in greater depth, from an experimental perspective. The left-hand side of Fig.\,\ref{Fig:Converge} shows the convergence of the empirical average of the samples (for each of the three considered pixels) towards its limit (the posterior mean\footnote{Considering that the convergence of the sampler to the posterior is provided (see \textsl{Convergence property} at the end of Sect.~\ref{Sec:Proposed}).\label{Foot:Converge}}). The figure shows effective convergence: as for the first pixel, convergence is essentially instantaneous, whereas for the other two pixels, it seems to take effect after approximately 300 iterations.

The right-hand side of Fig.\,\ref{Fig:Converge} shows the correlation of the chain for each of the three pixels (empirical correlation across iterations). For the first one, the correlation drops almost immediately to~$0$, which is consistent with the fact that the empirical average converges very quickly. For the other two pixels, it decreases less rapidly, and we can approximately consider that beyond 250 iterations, the correlation is slight. This decrease is also consistent with the convergence of the empirical average, which is slower than for the first pixel. 

\bfig[h]
\centering
\includegraphics[height=8cm]{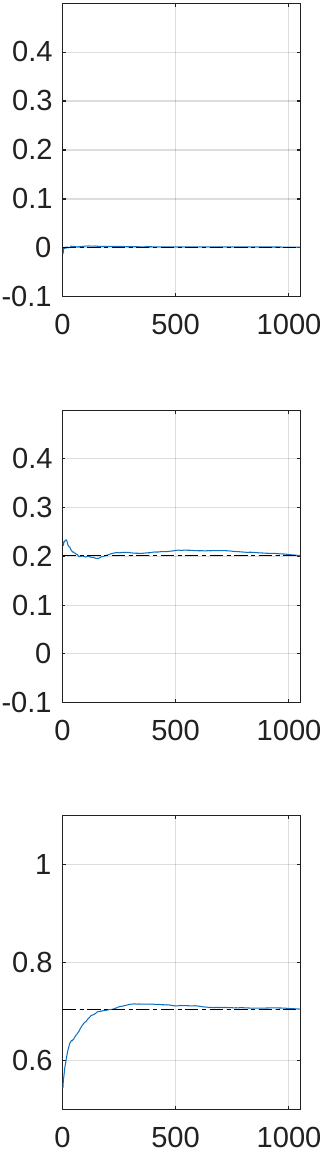}~~~~~\includegraphics[height=8cm]{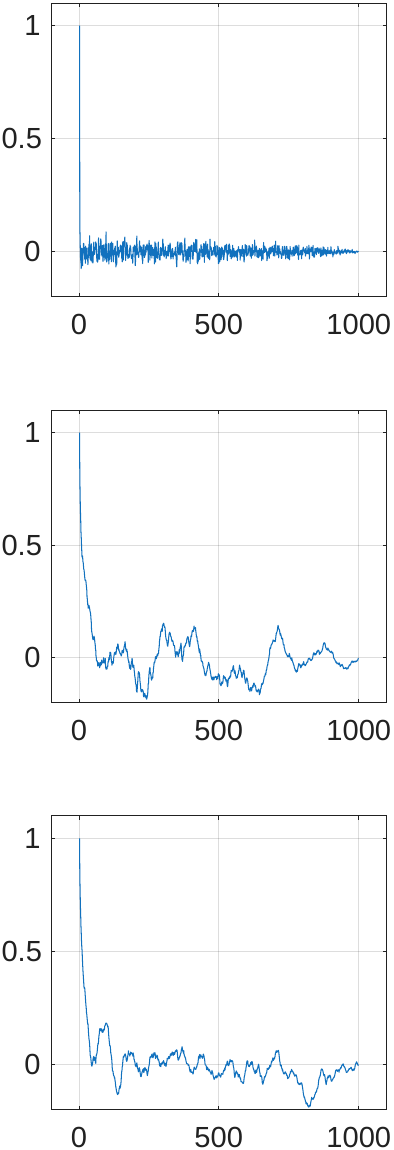}
\caption{Left part: empirical average of the samples provided by the Gibbs algorithm as a function of iteration index $n$ for three pixels of $\xb_{\sss 0}$. Right part: correlation of the chain (across the iterations of the Gibbs algorithm). Comments are given in Sect~\ref{Sec:SimulationsAdditional}.\label{Fig:Converge}}
\efig

\subsection{Efficiency, computation time and some comments} \label{Sec:SimulationsTime}


It should be noted that each iteration $n$ of the algorithm updates all the images (for $t=0,\dots,T$). Thus, the algorithm produces $N$ samples $\xb_{\sss 0:T}^{(n)}$ for $n=1,\dots N$ under the joint pdf for $\xb_{\sss 0:T}$. The images of interest $\xb_{\sss 0}^{(n)}$ are obtained by simple projective marginalisation, \ie discarding the $\xb_{\sss 1:T}^{(n)}$. 

As mentioned earlier, as iterations progress, the empirical average of the images $\xb_{\sss 0}^{(n)}$ (approximating the posterior mean\textsuperscript{\ref{Foot:Converge}}) is updated. The iterations stops when the difference between successive updates becomes smaller than a threshold, practically set to $10^{-2}$. The algorithm thus iterated $N=997$ times which took $53$~seconds. Most of the computation time (about 85\%) is due to the passage through the network. 

A particular feature of the G-DPS sampling scheme is that, ultimately, each iteration $n$ requires only a single pass through the neural network (to update $\xb_{\sss 0}$). Thus, scaling up to larger images does not appear to be a severe obstacle.


Moreover, the proposed G-DSP algorithm, as many others, could take advantage of parallelisation in the \Quote{batch} direction in order to simulate several independent samples in parallel. This possibility is supported by GPU architectures and most programming languages (Python, Matlab, etc.). We therefore have significant potential gains in terms of computing time.

Another major practical advantage of G-DSP is that it does not require the adjustement of any algorithm parameters or fine-tuning quantities  (apart from the threshold that puts an end to the iterations), unlike other algorithms, \eg~\cite{Chung24,Song23}.


A final point, by no means the least important. By its very nature, Gibbs’ algorithm (after burn-in period) is capable of generating a large number of samples, whereas existing algorithms restart the entire process in order to generate multiple samples. An additional advantage of G-DSP.



\section{Conclusion: G-DSP an attractive algorithm}

The paper deals with inverse problems when the observation model is linear with additive Gauss error and developments are within a Bayesian framework: it provides optimal estimate and uncertainty quantification. The prior relies on a diffusion model adjusted on a large set of examples. In that scenario, posterior sampling is known to be a challenging question and we propose a new answer. 

More precisely, we turn towards a Gibbs loop that splits the overall problem in far simpler sub-problems: iteratively sample each image under its conditional posterior. These conditionals are all Gaussian and their moments are easily calculated (using the backward pdf for the image of interest and the forward pdf for the latent variables). 
It seems that this idea has not been explored in the literature for this context (diffusion prior and inverse problem). This is possibly due to the fact that the community has focused on ideas inspired by the ancestral sampling with the intent to correct\,/\,guide its behaviour.

A crucial point is that of the convergence. Although a guarantee does not exists for G-DSP (since it partly relies on both forward and backward pdfs), there is a strong argument in favour of convergence in practice: the learning stage brings the forward and backward pdfs close together. In the limit case where the two pdfs are identical, convergence is guaranteed. 

The simulation study focuses on a toy example in image deconvolution based the MNIST example set. It shows that \linebreak[4] G-DPS  produces samples that provide  accurate image restoration, as well as precise and consistent elements for uncertainty quantification. This numerical study also confirms the remarkable computational efficiency of G-DSP.

Conclusively, the paper addresses the crucial question of sampling the posterior for inverse problem based on diffusion prior, offering an original, proper and highly efficient answer. 

To go further, it would be interesting to include the estimation of additional parameters \eg error level (hy\-per\-pa\-ra\-me\-ter estimation, automatic adjustment, self-tuned\dots)~\cite{Orieux12,Pereyra12} or instrument parameters~\cite{Orieux10,Campisi07,Kail12} (myopic, self-calibrated, semi-blind\dots). This question is partly addressed in~\cite{Giovannelli26a}. 
%

Two other lines of research appear promising. One involves seeking a joint distribution defined by the conditionals actually at work in G-DSP, and one might consider using a Hammersley-Clifford decomposition~\cite{Besag74}. The other involves developing an algorithm that would effectively be guaranteed to converge, for example a Metropolis-Hastings algorithm based on propositions derived from G-DSP.

Another perspective is to assess the G-DPS in comparison to other samplers, based for example on~\cite{Moroy26}.
\section*{Acknowledgment}
The author warmly thanks Liam Moroy, Guillaume Bourmaud, Fr\'ed\'eric Champagnat, Marcelo Pereyra and Charlesquin Kemajou for their help. Without them, I probably would not have entered this new world. A special thank you to Christelle Boillet for stimulating my creativity.

The work is conducted within project \emph{PEPR Origins} (ANR-22-EXOR-0016, supported by the France~2030 plan managed by Agence Nationale de la Recherche) and received financial support from the French government in the framework of the University of Bordeaux's France 2030 program \emph{RRI Origins}.

	\bibliographystyle{IEEEtran}
	\bibliography{biben,bibenabr,revuedef,revueabr,name,BaseBiblioJFG,BaseBiblioGPI,BaseBiblioAutre}

\begin{thebibliography}{10}
\providecommand{\url}[1]{#1}
\csname url@samestyle\endcsname
\providecommand{\newblock}{\relax}
\providecommand{\bibinfo}[2]{#2}
\providecommand{\BIBentrySTDinterwordspacing}{\spaceskip=0pt\relax}
\providecommand{\BIBentryALTinterwordstretchfactor}{4}
\providecommand{\BIBentryALTinterwordspacing}{\spaceskip=\fontdimen2\font plus
\BIBentryALTinterwordstretchfactor\fontdimen3\font minus
  \fontdimen4\font\relax}
\providecommand{\BIBforeignlanguage}[2]{{%
\expandafter\ifx\csname l@#1\endcsname\relax
\typeout{** WARNING: IEEEtran.bst: No hyphenation pattern has been}%
\typeout{** loaded for the language `#1'. Using the pattern for}%
\typeout{** the default language instead.}%
\else
\language=\csname l@#1\endcsname
\fi
#2}}
\providecommand{\BIBdecl}{\relax}
\BIBdecl

\bibitem{Giovannelli15}
J.-F. Giovannelli and J.~Idier, Eds., \emph{Regularization and {B}ayesian
  Methods for Inverse Problems in Signal and Image Processing}.\hskip 1em plus
  0.5em minus 0.4em\relax London: ISTE and John Wiley \& Sons Inc., 2015.

\bibitem{Kaipio05}
J.~Kaipio and E.~Somersalo, \emph{Statistical and computational inverse
  problems}.\hskip 1em plus 0.5em minus 0.4em\relax Berlin, Germany: Springer,
  2005.

\bibitem{Santamarina05}
J.~C. Santamarina and D.~Fratta, \emph{Discrete Signals and Inverse Problems:
  An Introduction for Engineers and Scientists}.\hskip 1em plus 0.5em minus
  0.4em\relax Chichester, England: WileyBlackwell, 2005.

\bibitem{Hansen10}
P.~C. Hansen, \emph{Discrete Inverse Problems: Insight and Algorithms}.\hskip
  1em plus 0.5em minus 0.4em\relax Philadelphia, USA: \uppercase{siam}, 2010.

\bibitem{Chan24}
S.~H. Chan, \emph{Tutorial on Diffusion Models for Imaging and Vision}, ser.
  Foundations and Trends in Machine Learning.\hskip 1em plus 0.5em minus
  0.4em\relax Hanover, \sca{ma}, USA: Now Publishers Inc, {J}an. 2024.

\bibitem{Ribeiro25}
F.~D.~S. Ribeiro and B.~Glocker, \emph{Demystifying Variational Diffusion
  Models}, ser. Foundations and Trends in Machine Learning.\hskip 1em plus
  0.5em minus 0.4em\relax Hanover, \sca{ma}, USA: Now Publishers Inc, {J}an.
  2025.

\bibitem{Nakkiran25}
P.~Nakkiran, A.~Bradley, and M.~Zhou, Hattieand~Advani, \emph{Step-by-Step
  Diffusion: An Elementary Tutorial}, ser. Foundations and Trends in Machine
  Learning.\hskip 1em plus 0.5em minus 0.4em\relax Hanover, \sca{ma}, USA: Now
  Publishers Inc, 2025.

\bibitem{Chung24}
H.~Chung, J.~Kim, M.~T. Mccann, M.~L. Klasky, and J.~C. Ye, ``Diffusion
  posterior sampling for general noisy inverse problems,''
  \emph{{I}nternational {C}onference on {L}earning {R}epresentations}, 2024.

\bibitem{Song23}
J.~Song, A.~Vahdat, M.~Mardani, and J.~Kautz, ``Pseudoinverse-guided diffusion
  models for inverse problems,'' in \emph{{I}nternational {C}onference on
  {L}earning {R}epresentations}, 2023.

\bibitem{Wu23}
L.~Wu, B.~Trippe, C.~Naesseth, D.~Blei, and J.~P. Cunningham, ``Practical and
  asymptotically exact conditional sampling in diffusion models,'' in
  \emph{{N}eu{RIPS}}, vol.~36, 2023, pp. 31\,372--31\,403.

\bibitem{Dou24}
Z.~Dou and Y.~Song, ``Diffusion posterior sampling for linear inverse problem
  solving: {A} filtering perspective,'' in \emph{{I}nternational {C}onference
  on {L}earning {R}epresentations}, {V}ienna, {A}ustria, {M}ay 2024.

\bibitem{Zhu24}
Y.~Zhu, Z.~Dou, H.~Zheng, Y.~Zhang, Y.~N. Wu, and R.~Gao, ``Think twice before
  you act: Improving inverse problem solving with {MCMC},'' 2024.

\bibitem{Cardoso23}
G.~Cardoso, Y.~Janati, S.~Le~Corff, and E.~Moulines, ``{M}onte {C}arlo guided
  diffusion for {B}ayesian linear inverse problems,'' 2023.

\bibitem{Coeurdoux24}
F.~Coeurdoux, N.~Dobigeon, and P.~Chainais, ``Plug-and-play split {G}ibbs
  sampler: Embedding deep generative priors in {B}ayesian inference,''
  \emph{\uppercase{ieee} {T}rans. {I}mage {P}rocessing}, vol.~33, pp.
  3496--3507, 2024.

\bibitem{Janati25}
\BIBentryALTinterwordspacing
Y.~Janati, E.~Moulines, J.~Olsson, and A.~Oliviero-Durmus, ``Bridging diffusion
  posterior sampling and {M}onte {C}arlo methods: a survey,''
  \emph{Philosophical Transactions of the Royal Society A}, vol. 383, no. 2299,
  2025. [Online]. Available: \url{https://doi.org/10.1098/rsta.2024.0331}
\BIBentrySTDinterwordspacing

\bibitem{Kamilov25}
N.~Yismaw, U.~S. Kamilov, and M.~S. Asif, ``Gaussian is all you need: A unified
  framework for solving inverse problems via diffusion posterior sampling,''
  \emph{\uppercase{ieee} {T}rans. {C}omputational {I}maging}, vol.~11, 2025.

\bibitem{Matlab23}
{MatlabDoc (fr.mathworks.com/help/)}, ``Generate images using diffusion,''
  deeplearning/ug/generate-images-using-diffusion.html, 2023.

\bibitem{Orieux12}
F.~Orieux, J.-F. Giovannelli, T.~Rodet, H.~Ayasso, M.~Husson, and A.~Abergel,
  ``Super-resolution in map-making based on a physical instrument model and
  regularized inversion. {A}pplication to {SPIRE/Herschel}.'' \emph{{A}stron.
  {A}strophys.}, vol. 539, {M}ar. 2012.

\bibitem{Pereyra12}
M.~Pereyra, N.~Dobigeon, H.~Batatia, and J.-Y. Tourneret, ``Segmentation of
  skin lesions in 2d and 3d ultrasound images using a spatially coherent
  generalized {R}ayleigh mixture model,'' \emph{\uppercase{ieee} {T}rans.
  {M}edical {I}maging}, vol.~31, no.~8, pp. 1509--1520, {A}ug. 2012.

\bibitem{Orieux10}
F.~Orieux, J.-F. Giovannelli, and T.~Rodet, ``{B}ayesian estimation of
  regularization and point spread function parameters for {W}iener--{H}unt
  deconvolution,'' \emph{{J}. {O}pt. {S}oc. {A}mer.}, vol.~27, no.~7, {J}uly
  2010.

\bibitem{Campisi07}
P.~Campisi and K.~Egiazarian, Eds., \emph{Blind Image Deconvolution}.\hskip 1em
  plus 0.5em minus 0.4em\relax CRC Press, 2007.

\bibitem{Kail12}
G.~Kail, J.-Y. Tourneret, F.~Hlawatsch, and N.~Dobigeon, ``Blind deconvolution
  of sparse pulse sequences under a minimum distance constraint: a partially
  collapsed {G}ibbs sampler method,'' \emph{\uppercase{ieee} {T}rans. {S}ignal
  {P}rocessing}, vol.~60, no.~6, pp. 2727--2743, {J}une 2012.

\bibitem{Giovannelli26a}
J.-F. Giovannelli, ``Estimation of instrument and noise parameters for inverse
  problem based on prior diffusion model,'' \emph{{P}roc. \uppercase{ieee}
  \uppercase{icip}}, 2026.

\bibitem{Besag74}
J.~E. Besag, ``Spatial interaction and the statistical analysis of lattice
  systems (with discussion),'' \emph{{J}. {R}. {S}tatist. {S}oc. {B}}, vol.~36,
  no.~2, pp. 192--236, 1974.

\bibitem{Moroy26}
L.~Moroy, G.~Bourmaud, F.~Champagnat, and J.-F. Giovannelli, ``On the posterior
  gap in plug \& play diffusion methods for sparse-view computed tomography,''
  \emph{IEEE Journal of Selected Topics in Signal Processing}, {J}an. 2026.

\end{thebibliography}

\edoc